\documentclass{article}
\usepackage{ijcai25}

\usepackage{times}
\usepackage{soul}
\usepackage{url}
\usepackage[hidelinks]{hyperref}
\usepackage[utf8]{inputenc}
\usepackage[small]{caption}
\usepackage{graphicx}
\usepackage{amsmath}
\usepackage{amsthm}
\usepackage{booktabs}
\usepackage{algorithm}
\usepackage{algorithmic}
\usepackage[switch]{lineno}

\usepackage{amssymb}
\usepackage{multirow}
\usepackage{colortbl}
\usepackage{xcolor}
\usepackage{array}
\usepackage{pifont}
\usepackage{bm}
\usepackage[misc]{ifsym}

\definecolor{mycolor}{RGB}{241,240,255}

\title{Point Cloud Mixture-of-Domain-Experts Model for 3D Self-supervised Learning}

\author{
Yaohua Zha$^{1,2}$
\and
Tao Dai$^3$ \thanks{Corresponding author.}\and
Hang Guo$^1$\and
Yanzi Wang$^1$\and
Bin Chen$^4$\and
Ke Chen$^2$\and
Shu-Tao Xia$^{2}$\\ 
\affiliations
$^1$Tsinghua Shenzhen International Graduate School, Tsinghua University\\
$^2$Institute of Perceptual Intelligence, Pengcheng Laboratory\\
$^3$College of Computer Science and Software Engineering, Shenzhen University\\
$^4$Harbin Institute of Technology, Shenzhen\\
}

\begin{document}

\maketitle

\begin{abstract}
Point clouds, as a primary representation of 3D data, can be categorized into scene domain point clouds and object domain point clouds. Point cloud self-supervised learning (SSL) has become a mainstream paradigm for learning 3D representations. However, existing point cloud SSL primarily focuses on learning domain-specific 3D representations within a single domain, neglecting the complementary nature of cross-domain knowledge, which limits the learning of 3D representations.
In this paper, we propose to learn a comprehensive \textbf{Point} cloud \textbf{M}ixture-\textbf{o}f-\textbf{D}omain-\textbf{E}xperts model (Point-MoDE) via a block-to-scene pre-training strategy. Specifically, 
We first propose a mixture-of-domain-expert model consisting of scene domain experts and multiple shared object domain experts.
Furthermore, we propose a block-to-scene pretraining strategy, which leverages the features of point blocks in the object domain to regress their initial positions in the scene domain through object-level block mask reconstruction and scene-level block position regression. By integrating the complementary knowledge between object and scene, this strategy simultaneously facilitates the learning of both object-domain and scene-domain representations, leading to a more comprehensive 3D representation.
Extensive experiments in downstream tasks demonstrate the superiority of our model.
\end{abstract}

\section{Introduction}

Recently, with the rapid development of 3D scanning technology, 3D point clouds have become the mainstream representation for 3D objects due to their ease of acquisition, explicit representation, and efficient storage. Point clouds can be categorized into scene domain point clouds~\cite{scannet,sunrgbd,s3dis} and object domain point clouds~\cite{modelnet,shapenet,scanobjectnn} based on the modeling object. As shown in Figure \ref{comp} (a), object domain point clouds describe specific objects or entities, such as an airplane, with relatively fewer points. Scene domain point clouds represent the entire environment or scene, such as indoor scenes, including multiple objects, structures, and background elements, with a larger number of points. Due to the significant disparity in point count and the elements being described, a substantial domain gap exists in these two types of point clouds.

\begin{figure}[t]
    \begin{center}
    \includegraphics[width=\linewidth]{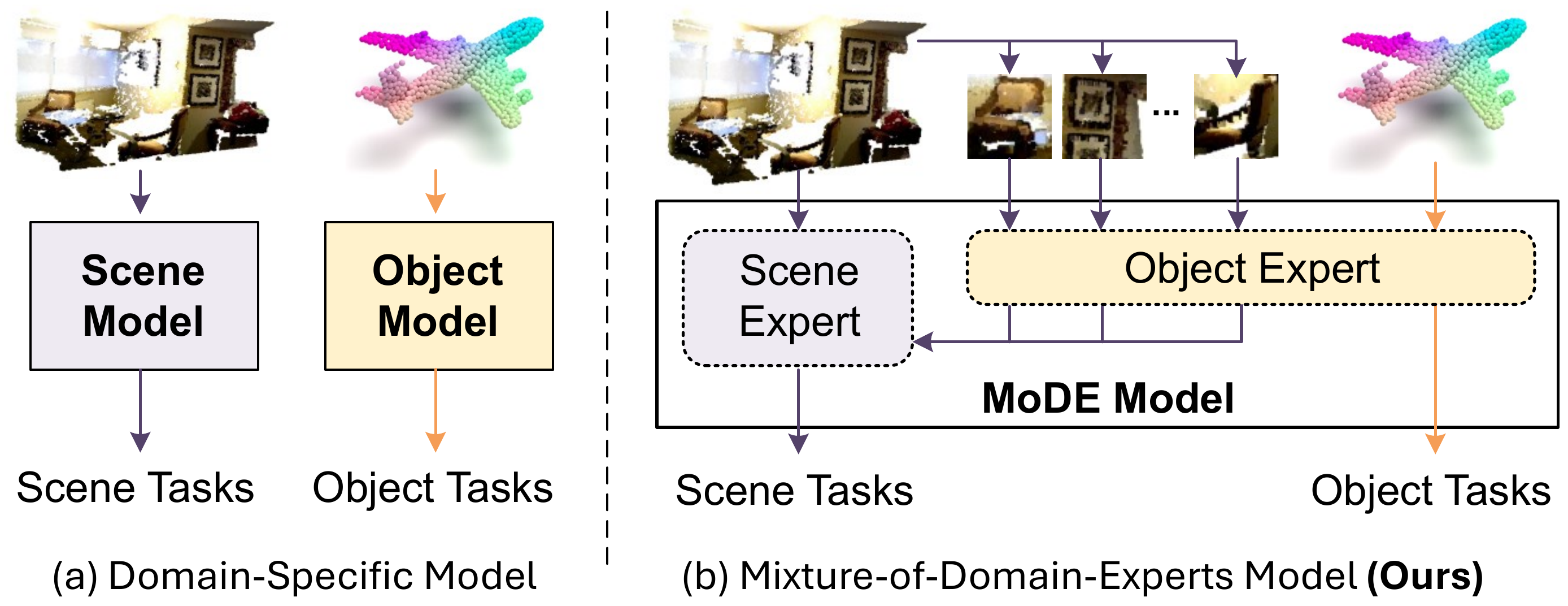}
    \caption{Comparison of (a) the traditional domain-specific model and (b) our mixture-of-domain-experts model for processing point clouds from different domains.
    }\label{comp}
    \end{center}
\end{figure}

Point cloud self-supervised learning (SSL)~\cite{pointbert,pointmae,m2ae,act}, pre-trained on massive point cloud data, have become the mainstream paradigm for learning 3D representations and have been widely transferred to various point cloud tasks~\cite{dapt,pma}. Despite the significant success, most of these methods are domain-specific due to the notable domain gap between object-domain and scene-domain point clouds. As shown in Figure \ref{comp} (a), these methods~\cite{pointbert,mae,m2ae} use scene-level data to learn scene model and object-level data for object model. This domain-independent learning approach not only fails to effectively integrate knowledge across different domains, but also struggles to perform well on data from multiple domains simultaneously. For example, Point-MAE~\cite{pointmae}, which is pre-trained on ShapeNet~\cite{shapenet}, primarily performs object point cloud tasks. For scene point clouds, it requires re-pretraining on scene-level datasets like ScanNet~\cite{scannet} to adapt to the scene domain. As shown in Figure \ref{comp} (c), directly transferring an object domain Point-MAE to scene tasks results in a significant performance drop. Similarly, models pre-trained on the scene domain also exhibit performance declines when transferred to object task.

\begin{figure}[t]
    \begin{center}
    \includegraphics[width=\linewidth]{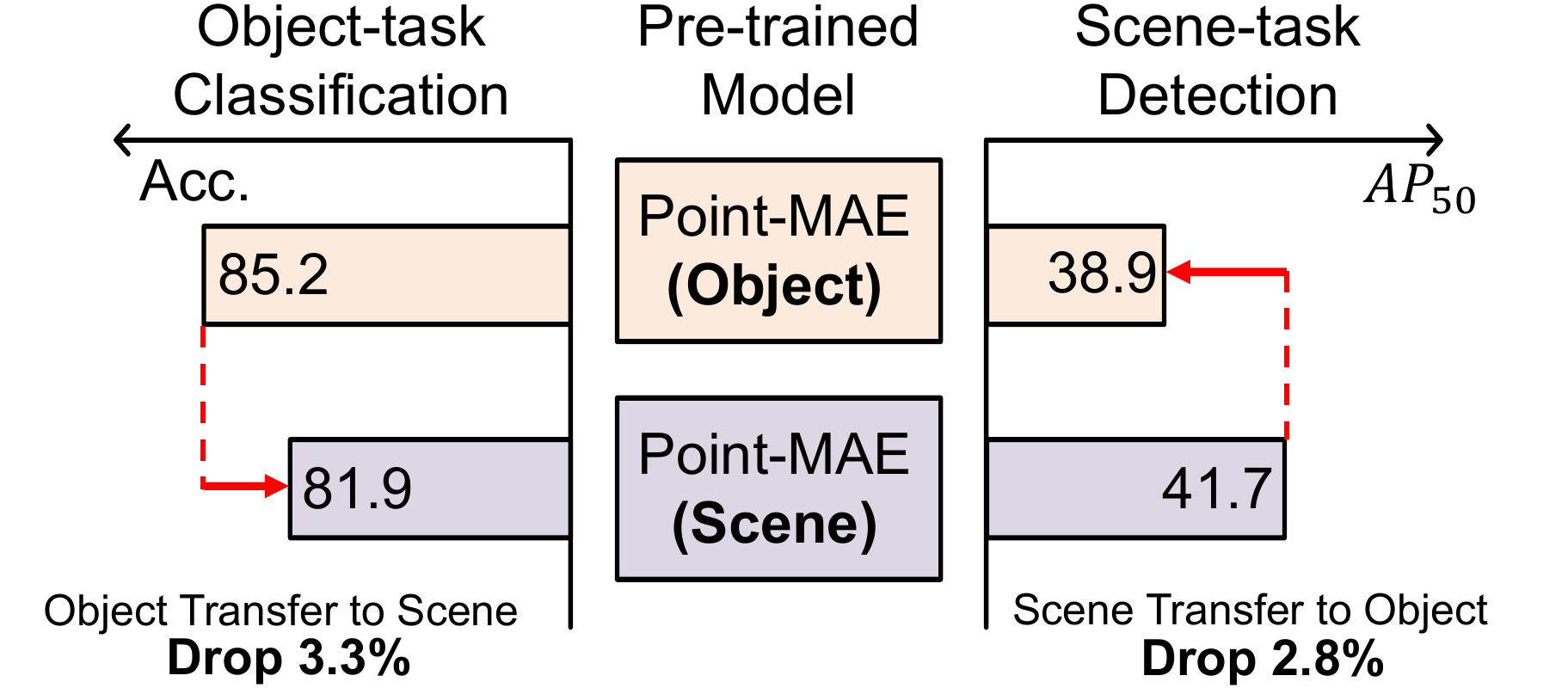}
    \caption{Transferability of the domain-specific Point-MAE model. Transferring a Point-MAE pre-trained on the scene to an object task results in a clear performance drop, and the same occurs for models in the object domain..
    }\label{comp}
    \end{center}
\end{figure}

Learning a more comprehensive 3D representation by simultaneously integrating knowledge from both the object domain and the scene domain is a reasonable solution. However, it is highly challenging for two main reasons. \textbf{Firstly}, the input data is inconsistent. Scene-level point clouds, such as ScanNet~\cite{scannet}, typically consist of 50k points, while object-level point clouds like ModelNet~\cite{modelnet} typically consist of 1k points. The disparity in point count between the two types is significant, making it difficult to process both types of data simultaneously using a single model. \textbf{Secondly}, there is inconsistency in task emphasis. Scene point clouds typically involve object detection or segmentation, which often prioritizes understanding fine-grained local point clouds. On the contrary, object point clouds generally involve classification tasks, which tend to prioritize understanding global geometry. 

To address the aforementioned challenges, we propose a block-to-scene pretraining strategy to pre-train a \textbf{Point} cloud \textbf{M}ixture-\textbf{o}f-\textbf{D}omain-\textbf{E}xperts model (Point-MoDE). 
We address the challenge of inconsistent input data by using domain-specific experts to process data from their respective domains. Additionally, we finetune the pre-trained model to address the inconsistency of task emphasis. Specifically, as shown in Figure \ref{comp} (b), we first design a point cloud mixture-of-domain-experts model that consists of a scene domain expert and multiple shared object domain experts. In the fine-tuning phase, for object domain data, our model selectively activates the object-domain expert for analysis. However, in the case of scene point clouds, we activate multiple shared object expert encoders to assist the scene expert in analyzing scene domain data collaboratively.

Furthermore, we propose a block-to-scene pre-training strategy that couples masked reconstruction and position regression tasks of random object blocks within a scene for self-supervised training, enabling us to train experts for different domains simultaneously. Specifically, we first randomly select point blocks within a scene and apply a set of transformations to convert each point block coordinates from the scene space to the object space. Then, within the object expert, we use a mask and reconstruction pipeline to recover the masked points of each block, enabling it to learn universal object representations. Finally, we introduce a scene-expert-based block position regression pipeline, which utilizes the blocks' features in the object space to regress these blocks' initial positions within the scene space, enabling the scene expert to learn scene representations with the assistance of the object experts. By block-to-scene pretraining, our Point-MoDE can simultaneously learn powerful object-level and scene-level representations and exhibit superior transferability. Our model can be fine-tuned directly on downstream tasks such as object point cloud classification, segmentation, completion, and scene point cloud detection without the need for any additional domain adaptation training. 

The main contributions can be summarized as follows:
\begin{itemize}
    \item  We propose a mixture-of-domain-experts model, integrating both scene domain and object domain experts to effectively process point clouds from different domains.
    \item We introduce a novel block-to-scene pretraining strategy that combines masked reconstruction and position regression tasks, enabling the simultaneous learning of object-level and scene-level representations.
    \item Extensive experiments across different datasets and tasks demonstrate the superiority and transferability of our model.
\end{itemize}

\begin{figure*}[h]
    \begin{center}
    \includegraphics[width=.85\linewidth]{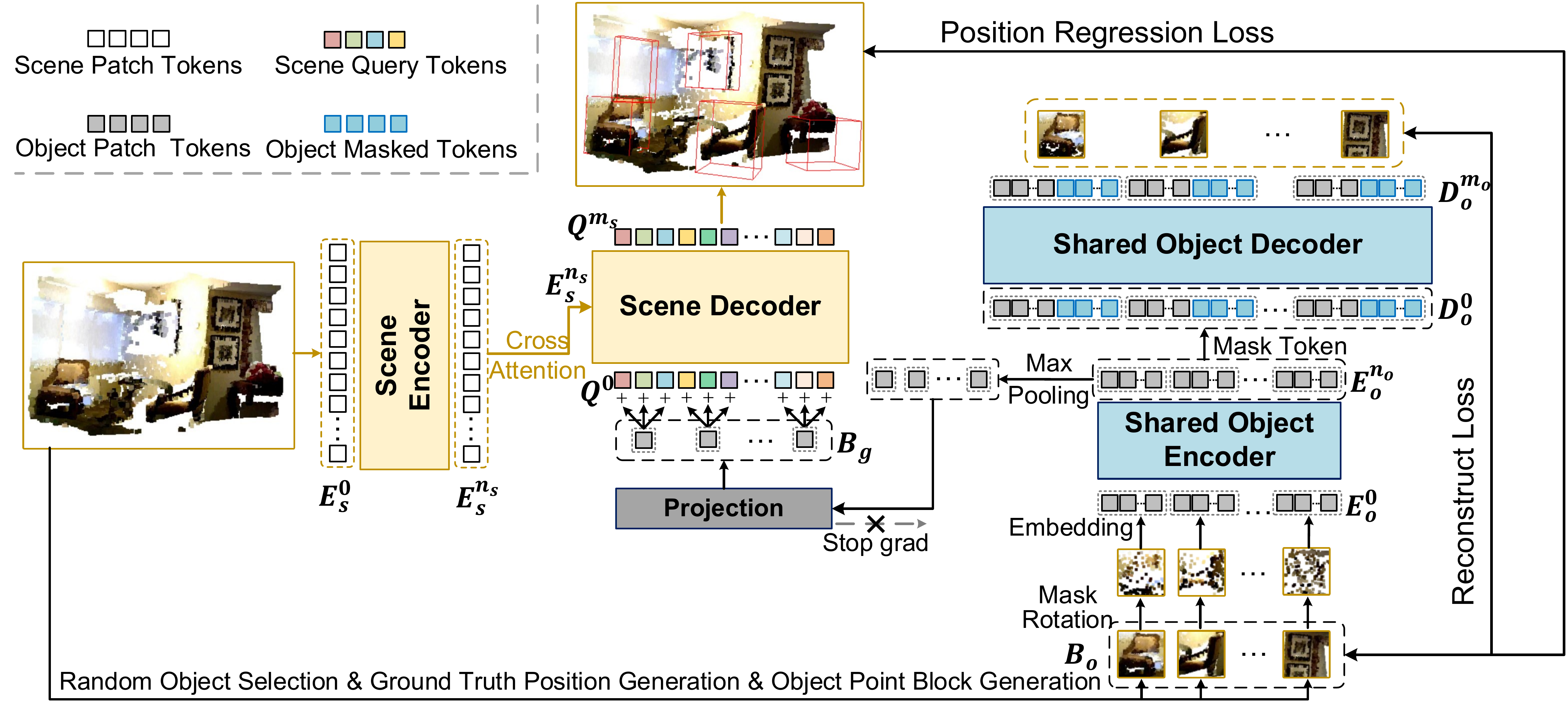}
    \caption{The architecture of our point cloud mixture-of-domain-experts model and the pipeline for block-to-scene pre-training. The left side illustrates the scene-level block position regression, while the right side shows the object-level block masked reconstruction.
    }\label{framework}
    \end{center}
\end{figure*}

\section{Related Work}



\subsection{Self-supervised Learning for Point Cloud}

Self-supervised learning, which enables the learning of general representations from large amounts of unlabeled data, has been widely applied in fields such as language~\cite{bert,gpt3,gpt4} and image~\cite{bao2021beit,simclr,chen2020generative,mae}. Inspired by the success of visual pretraining, numerous point cloud pretraining methods have also been proposed. Based on the pretext tasks, they can be categorized into contrastive learning paradigms~\cite{cpc,cmc} and masked reconstruction paradigms~\cite{bao2021beit,mae}.
PointContrast~\cite{pointcontrast}, CrossPoint~\cite{crosspoint}, and DepthContrast~\cite{depthcontrast} construct positive and negative sample pairs using various methods and employ contrastive learning techniques to learn 3D representations. 
Point-BERT~\cite{pointbert} was the first to propose learning universal 3D representations using the paradigm of masked reconstruction. Subsequently, numerous explorations have improved masked reconstruction from various perspectives. Point-MAE~\cite{pointmae} and Point-M2AE~\cite{m2ae} introduced the masked autoencoder for reconstruction, and PointGPT~\cite{pointgpt} proposed pretraining using an autoregressive approach. To address the limited amount of point cloud during pretraining, many approaches integrate multimodal knowledge to aid in learning point cloud features. ACT~\cite{act} leverages a pre-trained image model to assist in point cloud reconstruction, while I2P-MAE~\cite{i2pmae} employs an image-guided masking strategy. Despite the significant success, the above methods are domain-specific. In this paper, we propose training a mixture-of-domain-experts model to integrate knowledge from different domains for learning comprehensive 3D representations.

\section{Methodology}
In this section, we provide a detailed explanation of how to use our block-to-scene pretraining strategy to train our point cloud mixture-of-domain-experts model. 

\subsection{Point Cloud Mixture-of-Domain-Experts}

The overall architecture of our Point-MoDE is shown in Figure \ref{framework}, it is composed of four main components:  a scene expert encoder, a scene expert decoder, a shared object expert encoder, and a shared object expert decoder. It is primarily used for two main task pipelines: object point cloud processing and scene point cloud processing. 

Our Point-MoDE is a hybrid model. Due to significant differences across domains of point clouds and tasks, we selectively activate different sub-modules for various downstream data and tasks. For instance, in tasks such as object point cloud classification and object part segmentation, we selectively activate our object expert encoder according to specific tasks. For scene point cloud detection tasks, we activate all encoders. This collaborative approach is primarily adopted because utilizing the object expert encoder for analyzing local point blocks within the scene contributes to the scene encoder's comprehension of scene intricacies.

In the scene expert model, we adopt a standard Transformer as our scene expert, comprising 3 layers of Transformer-based encoders and 8 layers of Transformer-based decoders with a PointNet-based~\cite{pointnet} token embedding layer. In our object expert model, we directly leverage existing pre-trained object models based on mask-based point modeling, such as Point-BERT~\cite{pointbert}, Point-MAE~\cite{pointmae}, PointGPT~\cite{pointgpt}, etc. Using these models pre-trained on the object domain allows us to incorporate object-level priors into our pipeline while saving training resources and enabling better extension of existing models. This characteristic makes our object experts flexible, allowing us to replace them with different backbones.

\subsection{Block-to-Scene Pretraining}
Our block-to-scene pretraining primarily consists of the following three key components: random point block generation, object-level block masked reconstruction, and scene-level block position regression. Below, we provide a detailed illustration of the specific implementation of each component.

\subsubsection{Random Point Block Generation}

\textit{Random point block selection.} To leverage scene local details for scene understanding in an unsupervised manner, we randomly select $K_o$ local point blocks from the entire scene. We first random select $K_o$ points from the entire scene point cloud as the center points for each point block. For each center point, we then use the K-nearest neighbors algorithm to select the nearest $N_o$ points around it, forming the initial point block objects $\bm {B = \{B^1,...,B^{K_o}\}}$, where the $i$-th point block is $\bm {B^i}\in \mathbb{R}^{N_o\times 3}$. 

\textit{Ground-truth block position generation.} We generate the ground truth position of each random point block in the scene, which will be used to constrain the predicted position regressed by the final scene decoder. Inspired by the detection~\cite{detr,3detr,updetr} task, we use the 3D bounding box of each random point block as its ground truth position. By computing the mean of all points in each dimension of the entire point block, the coordinates of the center point are obtained. The half-lengths of the bounding box in each dimension are calculated by subtracting the minimum value from the maximum value in each dimension and dividing by 2. Subsequently, the center point coordinates and half-lengths in each dimension (x, y, and z) are concatenated to form the bounding box. Finally, standard procedures~\cite{3detr} are applied to compute bounding box parameters $\bm {B_b}$ such as size and corners for each bounding box.

\subsubsection{Object-level Block Masked Reconstruction}
We first initialize our object expert model using the Point-MAE (or Point-BERT, PointGPT, etc.) model pre-trained on ShapeNet~\cite{shapenet}. Since these models are open-source, we directly use the pre-trained weights provided by the official repository. This initialization allows us to incorporate object-level knowledge priors into our pipeline without the need for additional object data or specialized training.

\textit{Object point block generation.} We treat each randomly selected point block in $\bm B$ as an object point block and transform its coordinates from the scene space to the object space for processing by the object expert. Specifically, we apply a simple set of transformation functions to each point block. First, we subtract the coordinates of the center point from the $N_o$ local points to obtain the relative coordinates of each point. Then, we normalize these coordinates to the range [-1, 1]. Finally, after applying a random rotation transformation to each point block, we obtain all point blocks $\bm {B_o = \{B_o^1,...,B_o^{K_o}\}}$ as input to the object expert encoder. Through these transformation functions, we convert the coordinates of each point block from the scene space to the object space, decoupling the point block object coordinates from the original scene coordinates. This enables the object expert to learn the universal shape features of the point block objects.

\textit{Object point block masked and reconstruction.} 
We use a shared object autoencoder to perform mask-based reconstruction self-supervised learning on all generated object point blocks $\bm {B_o}$, enabling our object expert to learn a general representation of objects. We illustrate the entire mask and reconstruction process for the object point blocks using Point-MAE~\cite{pointmae} as an example. For the $i$-th object point block $\bm {B_o^i\in \mathbb{R}^{N_o\times 3}}$, we use farthest point sampling and the K-nearest neighbors algorithm to divide it into $M_o$ point patches. Then, after randomly masking most of the patches, we generate initial tokens and positional encodings for each unmasked patch using MLP-based token encoding and positional encoding. By adding them, we obtain the token $\bm {E}_{o}^{0}\in \mathbb{R}^{rM_o\times C_o}$ for each unmasked patch, where $r$ represents the unmasked ratio, and $C_o$ denotes the object feature dimension. Finally, we use a shared object expert encoder to extract object features $\bm {E}_{o}^{n_o}\in \mathbb{R}^{rM_o\times C_o}$, where $n_o$ is the number of layers in the scene expert.

In the decoding phase, we concatenate $\bm {E}_{o}^{n_o}$ with randomly initialized masked tokens to obtain $\bm {D}_{o}^{0}\in \mathbb{R}^{M_o\times C_o}$. Then, we use a standard Transformer-based decoder to decode, getting $\bm {D}_{o}^{m_o}\in \mathbb{R}^{M_o\times C_o}$. Finally, we use an MLP-based reconstruction head to reconstruct the coordinates of the masked point patches $\bm {R}_{o}^i\in \mathbb{R}^{N_o\times 3}$.

\subsubsection{Scene-level Block Position Regression}

\textit{Scene encoding.}
Given an input point cloud $\bm {P_s}\in \mathbb{R}^{N_s\times 3}$ with $N_s$ points, we first use farthest point sampling and the K-nearest neighbors algorithm to partition it into blocks. Then, using an MLP-based token encoding layer and a positional encoding layer, we generate the semantic token and positional encoding for each patch. By adding them, we obtain the token $\bm {E}_{s}^{0}\in \mathbb{R}^{M_s\times C_s}$ for each patch, where $M_s$ represents the number of scene patches, and $C_s$ denotes the scene feature dimension. Finally, we use a scene encoder based on the standard Transformer~\cite{attention} architecture to extract scene features $\bm {E}_{s}^{n_s}\in \mathbb{R}^{M_s\times C_s}$, where $n_s$ is the number of Transformer layers in the scene encoder.

\textit{Scene decoding and position regression.} We apply max pooling to the features of all blocks output by the object encoder to obtain the global feature for each block. After passing through the projection layer, these point block features are transformed into features $\bm {B}_{g}\in \mathbb{R}^{K_o\times C_s}$ for scene decoding input. However, we stop the gradients of $\bm {B}_{g}$ from propagating backward into the mask reconstruction pipeline during the backpropagation process, thereby mitigating the multi-task interference caused by the scene regression task on the object encoder. We will provide a detailed explanation of this issue in Section \ref{subsec:sg}. We then add the transformed point block features to randomly initialized queries to obtain enhanced queries $\bm {Q^0}\in \mathbb{R}^{q\times C_s}$, where $q$ is the number of queries. Since the number of point blocks and queries often differ, we replicate $\bm {B}_{g}$ to match all queries. 

We use a Transformer decoder based on self-attention and cross-attention as our scene expert decoder. The input queries $\bm {Q^0}$, with the assistance of the encoded features $\bm {E}_{s}^{n_s}$, pass through our decoder to obtain the decoded query features $\bm {Q^{m_s}}$, where $m_s$ is the number of scene expert decoder. Finally, we use different MLP-based reconstruction heads to predict the 3D bounding box $\bm {B_b^p}$ of each random point block.

\subsubsection{Loss Function.} 
We use a combination of mask reconstruction loss and point block regression loss to jointly constrain our pre-training process. For the reconstruction loss calculation, we follow previous work~\cite{pointmae,m2ae} and use Chamfer Distance~\cite{cdloss} (CD) as the loss function. For the regression loss calculation, we use Generalized Intersection over Union~\cite{giou} (GIoU) as the loss function. Therefore, our loss function is defined as follows:
\begin{gather}
    \label{eq1}
    \mathcal {L}=\lambda_1 \cdot {CD}(\bm {R_o},\bm {B_o})+ \lambda_2 \cdot {GIoU}(\bm {B_b^p},\bm {B_b})
\end{gather}
where $\lambda_1$ and $\lambda_2$ is a weighted combination of reconstruction loss and regression loss.

\begin{table*}[t]
  \centering
  \resizebox{0.85\textwidth}{!}{
    \begin{tabular}{lccllll}
    \toprule
    \multirow{2}[4]{*}{Method} & \multirow{2}[4]{*}{Reference}  & \multicolumn{1}{l}{\multirow{2}[4]{*}{\#Params (M)}}  & \multicolumn{3}{c}{ScanObjectNN} & \multirow{2}[4]{*}{ModelNet40} \\
    \cmidrule(lr){4-6}\textbf{}        &       &             & OBJ-BG & OBJ-ONLY & PB-T50-RS &  \\
    \midrule
    \multicolumn{7}{c}{\textit{Supervised Learning Only}} \\
    \midrule
    PointNet~\cite{pointnet} & CVPR 2017    & 3.5     & 73.3  & 79.2  & 68.0 & 89.2 \\
    PointNet++~\cite{pointnet++} & NeurIPS 2017    & 1.7   & 82.3  & 84.3  & 77.9  & 90.7 \\
    SFR~\cite{sfr} & ICASSP 2023     & -  & -     & -     & 87.8   & 93.9 \\
    PointMLP~\cite{pointmlp} & ICLR 2022     & 12.6  & -     & -     & 85.2   & 94.5 \\
    \midrule
    \multicolumn{7}{c}{\textit{Self-Supervised Learning}} \\
    \midrule
    Point-BERT~\cite{pointbert} & CVPR 2022   & 22.1  & 87.43 & 88.12 & 83.07 & 93.2 \\
    Point-MAE~\cite{pointmae} & ECCV 2022   & 22.1    & 90.02 & 88.29 & 85.18  & 93.8 \\
    Point-M2AE~\cite{m2ae} & NeurIPS 2022  & 15.3    & 91.22 & 88.81 & 86.43 & 94.0 \\
    PointGPT-S~\cite{pointgpt} & NeurIPS 2023   & 29.2   & 91.6 & 90.0 & 86.9  & 94.0 \\
    PointDif~\cite{pointdif} & CVPR 2024   & -   & 93.29 & 91.91 & 87.61  & -\\
    ACT\textcolor{red}{$^\dagger$}~\cite{act}   & ICLR 2023  & 22.1    & 93.29 & 91.91 & 88.21  & 93.7 \\
    LCM\textcolor{red}{$^\dagger$}~\cite{lcm} & NeurIPS 2024  & 2.7  & 94.51 & 92.75 & 88.87 & 94.2 \\
    Point-FEMAE\textcolor{red}{$^\dagger$}~\cite{femae} & AAAI 2024  & 27.4   & 95.18 & 93.29 & 90.22 & 94.5 \\
    \midrule
    Point-BERT\textcolor{red}{$^\dagger$} (baseline) & CVPR 2022   & 22.1  &  92.48 & 91.60 & 87.91 & 93.2 \\
    Point-MAE\textcolor{red}{$^\dagger$} (baseline) & ECCV 2022   & 22.1    &  92.67 &  92.08 & 88.27  & 93.8 \\
    PointGPT-B (baseline) & NeurIPS 2023  & 120.5  & 95.8 & 95.2 & 91.9 & 94.4 \\
    \midrule
    \rowcolor{mycolor} \textbf{Point-MoDE w/ Point-BERT\textcolor{red}{$^\dagger$}} & \textbf{Ours} & 22.1   & 93.46\textcolor{blue}{($\uparrow$ 1.0)} & 92.25\textcolor{blue}{($\uparrow$ 0.7)} & 88.58\textcolor{blue}{($\uparrow$ 0.7)}  & 93.6\textcolor{blue}{($\uparrow$ 0.4)} \\
    \rowcolor{mycolor}  \textbf{Point-MoDE w/ Point-MAE\textcolor{red}{$^\dagger$}} & \textbf{Ours} & 22.1   & 93.98\textcolor{blue}{($\uparrow$ 1.3)} & 92.77\textcolor{blue}{($\uparrow$ 0.7)} & 89.14\textcolor{blue}{($\uparrow$ 0.9)} & 94.0\textcolor{blue}{($\uparrow$ 0.2)} \\
    \rowcolor{mycolor} \textbf{Point-MoDE w/ PointGPT-B} & \textbf{Ours} & 120.5  & \textbf{96.9}\enspace\textcolor{blue}{($\uparrow$ 1.1)} & \textbf{95.6}\enspace\textcolor{blue}{($\uparrow$ 0.4)} & \textbf{92.4}\enspace\textcolor{blue}{($\uparrow$ 0.5)} & \textbf{94.7}\textcolor{blue}{($\uparrow$ 0.3)} \\
    \bottomrule
    \end{tabular}%
  }
  \caption{Classification accuracy on real-scanned (ScanObjectNN) and synthetic (ModelNet40) point clouds. In ScanObjectNN, we report the overall accuracy (\%) on three variants. In ModelNet40, we report the overall accuracy (\%) for both without and with voting. "\#Params" represents the model's parameter count. \textcolor{red}{$\dagger$} indicates that the pre-trained model used simple rotational augmentation during fine-tuning on ScanObjectNN.}
  \label{class}%
\end{table*}%

\section{Experiments}
\label{subsec:expall}
First, we pre-train the Point-MoDE model using our block-to-scene pretraining strategy based on point cloud data from the scene domain. After pre-training, we directly transfer the pre-trained model to various downstream tasks in different point cloud domains for fine-tuning. During fine-tuning, we selectively activate different sub-modules depending on the domain of the point cloud; for instance, we activate the object expert encoder for object point clouds, while utilizing all encoders for scene point clouds. This strategy enables Point-MoDE model, pre-trained with the block-to-scene approach, to outperform existing domain-specific models in most cases without any additional domain adaptation training, demonstrating the superiority and transferability of our model.
\subsection{Dataset and Pre-training}

We initialize our object expert model using the Point-MAE or other models pre-trained on ShapeNet~\cite{shapenet}, which can be directly obtained from these official repositories. Then, We combined all training data from the two most commonly used indoor scene datasets, SUNRGB-D~\cite{sunrgbd} and ScanNetV2~\cite{scannet}, to construct our block-to-scene pretraining dataset. Specifically, SUNRGB-D includes 5K single-view RGB-D training samples with oriented bounding box annotations for 37 object categories. 
ScanNetV2 contains 1.2K training samples, each with axis-aligned bounding box labels belonging to 18 object categories. We extracted 50K points for each of the 6.2K samples, using only the xyz coordinates of each point to construct the pretraining dataset.

During the pretraining phase, we input all 50K×3 point clouds into the scene encoder of Point-MoDE to extract scene-level features. Simultaneously, we randomly select 32 point blocks from the scene point cloud, with each block containing 2K local points, and input these into the 32 object encoders with shared parameters. We use the AdamW optimizer with a base learning rate of 5e-4 and a weight decay of 0.1. Simple rotation is applied as data augmentation to both the scene point cloud and each selected object point cloud. For the object point cloud masks, we set the mask ratio to 60\% following previous work~\cite{pointmae,act}. We train the model for 200 epochs using 8 A100 GPUs.

\subsection{Fine-tuning on Downstream Tasks}
\label{subsec:exp}

\subsubsection{Object Point Cloud Classification} 

We first evaluate the performance of our model on object point cloud classification tasks using the object expert encoder of Point-MoDE. We conduct point cloud classification on two of the most commonly used object point cloud datasets: ScanObjectNN~\cite{scanobjectnn} and ModelNet40~\cite{modelnet}. ScanObjectNN contains 15K real scanned point clouds, each with various backgrounds, occlusions, and noise, which effectively assesses the model's robustness. ModelNet40 includes 12K synthetic point clouds belonging to 40 different categories, with each point cloud being complete and clean, providing a better representation of 3D object shapes.

Following previous work~\cite{act,pointmamba}, we use 2K points as input for ScanObjectNN, apply simple rotation for data augmentation, and report classification accuracy without using voting. For ModelNet40, we use 1K points as input, apply scale and translate data augmentation, and report classification accuracy with the standard voting mechanism.

As presented in Table \ref{class}, \textbf{firstly}, compared to previous single-domain baselines (Point-MAE, Point-BERT, and PointGPT), our Point-MoDE method achieves consistent and stable improvements, indicating that leveraging scene domain knowledge to assist object domain learning is an effective strategy. This approach not only preserves object representations but also enables the learning of more comprehensive knowledge. \textbf{Secondly}, our Point-MoDE shows a significant improvement on the real-world scanned point cloud dataset ScanObjectNN, while the improvement on the synthetic point cloud dataset ModelNet40 is smaller. This is because we use randomly sampled scene data during pretraining, which lacks explicit semantics. Such non-semantic data is more similar to real scanned point clouds, thereby aiding the model in learning more realistic point cloud representations.

\subsubsection{Object Point Cloud Part Segmentation}  We assess the performance of our Point-MoDE in part segmentation using the ShapeNetPart dataset~\cite{shapenet}, comprising 16,881 samples across 16 categories. 
We utilize the same segmentation setting after the pre-trained encoder as in previous works \cite{pointmae,m2ae} for fair comparison. 
The experimental results displayed in Table \ref{seg}. Our method also achieves consistent improvements compared to single-domain approaches, demonstrating the superiority of leveraging object knowledge to assist in scene tasks. This improvement appears less pronounced compared to detection tasks, which can be attributed to the inherently challenging nature of fine-grained tasks like segmentation. Therefore, even relatively small improvements are sufficient to highlight the effectiveness of our approach.

\begin{table}[t]
  \centering
  \resizebox{\linewidth}{!}{
    \begin{tabular}{lcll}
    \toprule
    Methods & Reference & $\mathrm{mIoU}_{c}$ & $\mathrm{mIoU}_{I}$ \\
    \midrule
    \multicolumn{4}{c}{\textit{Supervised Learning Only}} \\
    \midrule
    PointNet++~\cite{pointnet++}  & NeurIPS 2017& 81.9  & 85.1 \\
    PointMLP~\cite{pointmlp}  & ICLR 2022 & 84.6  & 86.1 \\
    \midrule
    \multicolumn{4}{c}{\textit{Single-Modal Self-Supervised Learning}} \\
    \midrule
    PointContrast~\cite{pointcontrast}  & ECCV 2020 & -  & 85.1 \\
    IDPT~\cite{idpt}  & ICCV 2023 & 83.8  & 85.9 \\
    Point-BERT~\cite{pointbert}  & CVPR 2022 & 84.1  & 85.6 \\
    MaskPoint~\cite{maskpoint}  & ECCV 2022 & 84.4  & 86.0 \\
    Point-MAE~\cite{pointmae}  & ECCV 2022 & 84.2  & 86.1 \\
    ACT~\cite{act}  & ICLR 2023 & 84.7  & 86.1 \\
    Point-FEMAE~\cite{femae} & AAAI 2024& 84.9  & 86.3 \\
    PointGPT-B~\cite{pointgpt}  & NeurIPS 2023 & 84.5  & 86.5 \\
    \rowcolor{mycolor} \textbf{Point-MoDE w/ Point-BERT} & Ours & 84.6\textcolor{blue}{($\uparrow$ 0.5)} & 86.2\textcolor{blue}{($\uparrow$ 0.6)} \\
    \rowcolor{mycolor} \textbf{Point-MoDE w/ Point-MAE} & Ours & 84.5\textcolor{blue}{($\uparrow$ 0.3)} & 86.3\textcolor{blue}{($\uparrow$ 0.2)} \\
    \rowcolor{mycolor} \textbf{Point-MoDE w/ PointGPT-B} & Ours & \textbf{84.9}\textcolor{blue}{($\uparrow$ 0.4)} & \textbf{86.5}\textcolor{gray}{($-$)} \\
    \bottomrule
    \end{tabular}%
    }
  \caption{Part segmentation results on the ShapeNetPart. The mean IoU across all categories, i.e., $\mathrm{mIoU}_{c}$ (\%), and the mean IoU across all instances, i.e., $\mathrm{mIoU}_{I}$ (\%) are reported.}
  \label{seg}%
\end{table}%

\subsubsection{Scene Point Cloud Detection} 
We further fine-tune the pre-trained Point-MoDE on scene-level object detection tasks. At this stage, we primarily rely on the scene expert encoder to process scene-level inputs. Simultaneously, we randomly select 32 point blocks from the scene point cloud and use the 32 shared object encoders to handle these local point blocks. During the decoding phase, we integrate the results from the scene expert encoder with random queries from the scene, helping the scene expert to better understand scene details. We use ScanNetV2~\cite{scannet}, to evaluate our model's scene understanding capabilities. 

Table \ref{dete} shows our detection results, our Point-MoDE model has shown significant improvements compared with other single domain models. In particular, there is an improvement of 6-9\% in the $AP_{50}$ metric. This substantial improvement is mainly attributed to our Point-MoDE using multiple object encoders to assist the scene encoder in analyzing the overall scene. This approach enables the scene model to focus on more local details, thereby enhancing scene understanding. 

\begin{table}[t]
  \centering
  \resizebox{\linewidth}{!}{
    \begin{tabular}{lcll}
    \toprule
     Methods    & Reference   & $AP_{25}$  & $AP_{50}$ \\
     \midrule
     \multicolumn{4}{c}{\textit{Supervised Learning Only}} \\
     \midrule
     VoteNet~\cite{votenet} & ICCV 2019 & 58.6  & 33.5 \\
     3DETR~\cite{3detr}  & ICCV 2021 & 62.1  & 37.9 \\
     \midrule
     \multicolumn{4}{c}{\textit{Self-Supervised Learning}} \\
     \midrule
     PointContrast~\cite{pointcontrast}  & ECCV 2020 & 58.5  & 38.0 \\
     DepthContrast~\cite{depthcontrast} & ICCV 2021 & 61.3  & - \\
     Point-BERT~\cite{pointbert}  & CVPR 2022 & 61.0  & 38.3 \\
     PiMAE~\cite{pimae}  & CVPR 2023 & 62.6  & 39.4 \\
     Point-MAE~\cite{pointmae}  & ECCV 2022 & 59.5  & 41.2 \\
     ACT~\cite{act}    & ICLR 2023 & 63.8  & 42.1 \\
     MaskPoint~\cite{maskpoint} & ECCV 2022 & 64.2  & 42.1 \\
     PointGPT-B~\cite{pointmae}  & NeurIPS 2023 & 63.1  & 42.8 \\
     PointDif~\cite{pointdif}  & CVPR 2024 & -  & 43.7 \\
     \rowcolor{mycolor} \textbf{Point-MoDE w/ Point-BERT} & Ours & 65.4\textcolor{blue}{($\uparrow$ 4.4)} & 47.1\textcolor{blue}{($\uparrow$ 8.8)} \\
     \rowcolor{mycolor} \textbf{Point-MoDE w/ Point-MAE} & Ours & 66.2\textcolor{blue}{($\uparrow$ 6.7)} & 48.8\textcolor{blue}{($\uparrow$ 7.6)} \\
     \rowcolor{mycolor} \textbf{Point-MoDE w/ PointGPT-B} & Ours & \textbf{66.7}\textcolor{blue}{($\uparrow$ 3.6)} & \textbf{48.9}\textcolor{blue}{($\uparrow$ 6.1)} \\
    \bottomrule
    \end{tabular}%
    }
  \caption{Object detection results on ScanNetV2. We adopt the average precision with 3D IoU thresholds of 0.25 ($AP_{25}$) and 0.5 ($AP_{50}$) for the evaluation metrics.}
  \label{dete}%
\end{table}%

\subsection{Scene Semantic Segmentation} 

We have conducted experiments on scene-level semantic segmentation tasks to assess the performance of Point-MoDE in classifying each point in a scene into semantic categories. We validated our model using the indoor S3DIS~\cite{s3dis} dataset for semantic segmentation tasks. Specifically, we tested the model on Area 5 while training on other areas and report the mean IoU (mIoU) and mean Accuracy (mAcc). To ensure a fair comparison, we used the same codebase based on the PointNext~\cite{pointnext} baseline and employed identical decoders and semantic segmentation heads. 

The experimental results are shown in the table \ref{semseg}. Compared to training the PointNeXt model from scratch, our method improves the mIoU score by 2.3\%. It also shows significant improvements over other pretraining models, such as Point-MAE~\cite{pointmae} and PointDif~\cite{pointdif}. This enhancement is largely due to our block-to-scene pretraining, which equips the model with strong scene understanding capabilities and further demonstrates the generalizability of our approach.

\begin{table}[htbp]
  \centering
  \resizebox{\linewidth}{!}{
    \begin{tabular}{lcll}
    \toprule
    Methods & Reference & mIoU  & mAcc \\
    \midrule
    \multicolumn{4}{c}{\textit{Supervised Learning Only}} \\
    \midrule
    PointNeXt~\cite{pointnext} & NeurIPS 2022 & 68.5  & 75.1 \\
    Pix4Point~\cite{pix4point} & 3DV 2024  & 69.6  & 75.2 \\
    \midrule
    \multicolumn{4}{c}{\textit{Self-Supervised Learning}} \\
    \midrule
    Point-BERT~\cite{pointbert} & CVPR 2022 & 68.9  & 76.1 \\
    MaskPoint~\cite{maskpoint} & ECCV 2022 & 68.6  & 74.2 \\
    Point-MAE~\cite{pointmae} & ECCV 2022 & 68.4  & 76.2 \\
    PointDif~\cite{pointdif}  & CVPR 2024 & 70.0    & 77.1 \\
    PointGPT-B~\cite{pointmae}  & NeurIPS 2023 & 70.4  & 77.9 \\
    \rowcolor{mycolor} \textbf{Point-MoDE} w/ Point-BERT & Ours & 69.3\textcolor{blue}{($\uparrow$ 0.4)}  & 76.8\textcolor{blue}{($\uparrow$ 0.7)} \\
    \rowcolor{mycolor} \textbf{Point-MoDE} w/ Point-MAE & Ours & 69.2\textcolor{blue}{($\uparrow$ 0.8)}  & 76.6\textcolor{blue}{($\uparrow$ 0.4)} \\
    \rowcolor{mycolor} \textbf{Point-MoDE} w/ PointGPT-B & Ours & 70.9\textcolor{blue}{($\uparrow$ 0.5)}  & 78.1\textcolor{blue}{($\uparrow$ 0.2)} \\
    \bottomrule
    \end{tabular}%
  }
  \caption{Semantic segmentation results on S3DIS Area 5.}
  \label{semseg}%
\end{table}%

\subsection{Ablation Study}

\subsubsection{The Impact of Different Pretraining Strategies} 

\begin{table}[htbp]
  \centering
  \resizebox{\linewidth}{!}{
    \begin{tabular}{cccccc}
    \toprule
    Scene B. R. & Object M. R. & \multicolumn{1}{l}{Coordinate S. T.} & Joint T. & Classification & Detection \\
    \midrule
    \ding{52} & \ding{56} & \ding{56} & \ding{56} & 88.27 & 45.8 \\
    \midrule
    \ding{56} & \ding{52} & \ding{56} & \ding{56} & 88.32 & 41.2 \\
    \ding{56} & \ding{52} & \ding{52} & \ding{56} & 88.86 & 41.2 \\
    \midrule
    \ding{52} & \ding{52} & \ding{56} & \ding{56} & 88.32 & 43.8 \\
    \ding{52} & \ding{52} & \ding{52} & \ding{56} & 88.86 & 43.8 \\
    \midrule
    \ding{52} & \ding{52} & \ding{56} & \ding{52} & 87.21 & 47.4 \\
    \ding{52} & \ding{52} & \ding{52} & \ding{52} & 89.14 & 48.8 \\
    \bottomrule
    \end{tabular}%
  }
  \caption{The Impact of Different Pretraining Strategies.}
  \label{stra}%
\end{table}%

We first analyze the impact of each key component in the block-to-scene pretraining strategy. Our block-to-scene pretraining strategy consists of four key parts: scene-level block position regression (Scene B. P.), object-level block mask reconstruction (Object M. R.), coordinate space transformation (Coordinate S. T.) of object blocks (from scene space to object space), and joint pertaining (Joint T.) of objects and scenes (connecting objects and scenes via the Projection module in Figure 3). A detailed analysis of these components is provided in Table \ref{stra}.

When using only Scene B. P. or Object M. R., the pretraining strategy enhances only the subsequent scene tasks or object tasks, failing to achieve simultaneous improvements. Even when both strategies are employed without Joint T., while simultaneous enhancement is possible, the inability of the two pipelines to interact prevents the complementary knowledge from being effectively learned, thereby limiting the representation capability. Only by jointly employing both regression and reconstruction can the most comprehensive representation be achieved.

Secondly, Coordinate S. T. is crucial for learning object representations. This is because the coordinates of blocks directly cropped from the scene belong to the scene space and are often not normalized within the range of [-1, 1]. In contrast, most object-level point cloud analysis assumes coordinates within the range of [-1, 1]. Thus, transforming the coordinates is essential. As shown in the table, omitting this transformation results in performance degradation on object tasks. This is because the scene-space coordinates disrupt the learning of object representations, highlighting the importance of this normalization step for effective object-level analysis.

\subsubsection{The Impact of Stop Gradients} 
\label{subsec:sg}

\begin{table}[htbp]
  \centering
  \resizebox{\linewidth}{!}{
    \begin{tabular}{lrccc}
    \toprule
    \multicolumn{1}{c}{\multirow{2}[4]{*}{Methods}} & \multicolumn{2}{c}{Object Classification} & Scene Detection &  \\
\cmidrule{2-5}          & \multicolumn{1}{l}{ModelNet40} & ScanObjectNN & AP25  & AP50 \\
    \midrule
    w/o S. G. & 93.4  & 87.45 & 62.7  & 41.5 \\
    w/ S. G.  & 94.0  & 89.14 & 66.2  & 48.8 \\
    \bottomrule
    \end{tabular}%
  }
  \caption{The impact of stop gradients.}
  \label{sg}%
\end{table}%

In our implementation, gradient stopping plays a crucial role. This mechanism allows different encoders within the model to jointly learn domain-specific representations during block-to-scene pretraining without interfering with each other. Thus, ensuring the accuracy and independence of each encoder's learning objectives during training is of paramount importance.

Since the position regression objective and the masked reconstruction objective are two distinct tasks in our pretraining process, failing to decouple the learning processes of the different encoders could lead to catastrophic forgetting. For instance, without gradient stopping, gradients from the object-level reconstruction tasks could backpropagate into the scene's encoder, interfering with its ability to learn scene-level knowledge. By applying gradient stopping, we effectively prevent this interference, ensuring that each encoder remains focused on its specific task and thereby avoiding catastrophic forgetting. Table \ref{sg} reports the impact of applying gradient stopping. When the gradient stopping mechanism is removed, the AP50 on scene-level detection tasks drops by 7.3\%, clearly indicating that the learning process of the scene encoder is disrupted. Similarly, significant performance degradation is observed in classification tasks. These results collectively demonstrate that decoupling the representation learning of different encoders is essential for preventing catastrophic forgetting.

\subsubsection{The Impact of Object Expert Initialization}

\begin{table}[htbp]
  \centering
  \resizebox{\linewidth}{!}{
    \begin{tabular}{lrccc}
    \toprule
    \multicolumn{1}{c}{\multirow{2}[4]{*}{Methods}} & \multicolumn{2}{c}{Object Classification} & Scene Detection &  \\
\cmidrule{2-5} & \multicolumn{1}{l}{ModelNet40} & ScanObjectNN & AP25  & AP50 \\
    \midrule
    Scratch & 93.2  & 88.38 & 65.7  & 48.5 \\
    Pre-trained Point-MAE & 94.0    & 89.14 & 66.2  & 48.8 \\
    \bottomrule
    \end{tabular}%
  }
  \caption{The impact of object expert initialization.}
  \label{init}%
\end{table}%

We further analyze the impact of initializing the object expert model. In our implementation, we initialize the model using an object model pre-trained on ShapeNet, after which no object data is used. This initialization ensures that we can learn representations from the object domain prior. We further examine the effect of not using object priors (Scratch), where no object domain data is used during training. The results, as shown in Table \ref{init}, indicate that the absence of object priors has a minimal impact on scene representation learning, resulting in only slight performance degradation. However, it significantly affects the learning of object point cloud representations. 
For the real point cloud ScanObjectNN, even when trained from scratch, the model eventually surpasses the one trained only on ShapeNet55 (88.27). However, for the synthetic point cloud ModelNet40, there is a larger gap compared to the model trained only on ShapeNet55 (93.8). This is because the point blocks cut from the scene lack specific semantics, making their distribution more similar to real point clouds but significantly different from the distribution of synthetic point clouds, leading to poorer performance on synthetic point clouds.

\section{Conclusion}

In this paper, we propose a block-to-scene pretraining strategy to train a mixture-of-domain-experts model that seamlessly integrates scene domain and object domain experts. This strategy enables the simultaneous acquisition of object-level and scene-level representations, enabling the learning of comprehensive 3D representations. Extensive experiments conducted across various datasets and tasks validate the superiority and transferability of our proposed approach. Our findings highlight the potential of leveraging cross-domain knowledge to achieve robust and versatile 3D representation learning, paving the way for future study.

\section*{Acknowledgements}

This work is supported in part by the National Natural Science Foundation of China, under Grant (62302309,62171248), Shenzhen Science and Technology Program (JCYJ20220818101014030,JCYJ20220818101012025), and the PCNL KEY project (PCL2023AS6-1).

\bibliographystyle{named}
\bibliography{ijcai25}

\end{document}